\documentclass[12pt,longbibliography,eprint]{revtex4-1}
\usepackage{epigraph}
\usepackage{graphicx}
\usepackage{epstopdf}
\usepackage{hyperref}

\begin{document}
\preprint{V.M.}
\title{Multiple--Instance Learning: Christoffel Function Approach to Distribution Regression Problem.
}
\author{Vladislav Gennadievich \surname{Malyshkin}} 
\email{malyshki@ton.ioffe.ru}
\affiliation{Ioffe Institute, Politekhnicheskaya 26, St Petersburg, 194021, Russia}

\date{November, 19, 2015}

\begin{abstract}
\begin{verbatim}
$Id: DistReg.tex,v 1.53 2015/11/22 23:16:03 mal Exp $
\end{verbatim}
A two--step Christoffel function based solution is proposed
to distribution regression problem.
On the first step,
to model distribution of observations inside a bag,
build Christoffel function for each bag of observations.
Then, on the second step,
build outcome variable Christoffel function,
but use the bag's Christoffel function value at given point
as the weight for the bag's outcome.
The approach allows the result to be obtained in closed
form and then to be  evaluated numerically.
While most of existing  approaches minimize
some kind an error between outcome and prediction,
the proposed approach is conceptually different,
because it uses Christoffel function
for knowledge representation,
what is conceptually equivalent working with probabilities only.
To receive possible outcomes and their probabilities
Gauss quadrature for second--step measure can be built,
then the nodes give possible outcomes and
normalized weights -- outcome probabilities.
A library providing numerically stable polynomial basis
for these calculations is available, what make the proposed approach practical.
\end{abstract}

\keywords{Distribution Regression, Christoffel Function}
\maketitle

\section{\label{intro}Introduction}
Multiple instance learning is an important Machine Learning (ML)
concept having numerous applications\cite{yang2005review}.
In multiple instance learning class label is associated not with a
single observation, but with a ``bag'' of observations.
A very close problem is distribution regression
problem, where a $l$-th sample distribution of $x$
is mapped to a single $y^{(l)}$ value.
There are numerous heuristics methods developed from both: ML and distribution regression sides, see \cite{szabo2014learning} for review.
As in any ML problem the most important part is not so much
the learning algorithm, but the way how the learned knowledge is represented.
Learned knowledge is often represented as a set of propositional rules,
regression function,
Neural Network weights, etc. Most of the approaches minimize
an error between result  and prediction,
and some kind of $L^2$ metric is often used as an error.
The simplest example of such approach is least squares--type  approximation.
However, there is exist different kind of approximation,
Radon--Nikodym type, that operates not with result error,
but with sample probabilty, see the Ref. \cite{2015arXiv151101887G}
as an example comparing two these approaches.

Similar transition from result error to
probability of outcomes is made in this paper.
In this work we use Christoffel function
as a mean to store knowledge learned.
Christoffel function
is a very fundamental concept related
to ``distribution density'', quadratures weights, number of observations, etc\cite{totik,nevai}.
Recent progress in numerical stability
of high order distribution moments calculation\cite{2015arXiv151005510G}
allows Christoffel function to be built to a very high order,
what make practical the approach of using Christoffel function
as way to represent knowledge learned.

The paper is organized as following:
In Section \ref{christoffelfun}
a general theory of distribution regression is discussed and
close form result, Eq. (\ref{wxy}), is presented.
Then in Section \ref{christoffelnum} numerical example of
Eq. (\ref{wxy}) application is presented. In Section \ref{christoffeldisc}
possible further  development is discussed.

\section{\label{christoffelfun}Christoffel Function Approach}
Consider distribution regression problem
where a bag of $N$ observations $x$
is mapped to a single outcome observation $y$ for $l=[1..M]$.
\begin{eqnarray}
  (x_1,x_2,\dots,x_j,\dots,x_N)^{(l)}&\to&y^{(l)}  \label{regressionproblem}
\end{eqnarray}
A distribution regression problem can have a goal to estimate
$y$, average of $y$, distribution of $y$, etc.
given specific value of $x$. While the
Christoffel function can be used as a proxy to probabilty
estimation,  but for ``true'' distribution estimation a complete
Gauss quadrature should be built, then the nodes
would give possible outcomes and normalized weights -- outcome probabilities.

For further development we need $x$ and $y$ bases $Q_k(x)$ and $Q_m(y)$
and some $x$ and $y$ measure.
For simplicity, not reducing the generality of the approach,
we are going to assume that $x$ measure is a
sum over $j$ index  $\sum_j$,  $y$ measure is a $\sum_l$,
the basis functions $Q_k(x)$ are polynomials $k=0..d_x-1$, and
$Q_m(y)$  are polynomials $m=0..d_y-1$ where $d_x$ and $d_y$ is the
number of elements in $x$ and $y$ bases, typical value for $d_x$ and $d_y$ is below 10--15.

If no $x$ observations exist in each bag ($N=0$), how to estimate the number of observations
for given $y$ value? The answer is Christoffel function $\lambda(y)$.
\begin{eqnarray}
  G_y&=&<Q_sQ_t>_y=\sum_{l=1}^{M} Q_s(y^{(l)})Q_t(y^{(l)}) \label{QQy} \\
  K(z,y)&=&\sum_{s,t=0}^{d_y-1}Q_s(z)\left(G_y\right)^{-1}_{st}Q_t(y) \label{Ky} \\
  \lambda(y)&=&\frac{1}{K(y,y)} \label{w}  
\end{eqnarray}
The $G_y$ is Gramm matrix,
$K(y,y)$ is a positive quadratic form with matrix
equal to Gramm matrix inverse and
is a polynomial of $2d_y-2$ order, when the form is expanded.
The $K(z,y)$ is a reproducing kernel: $P(z)=\sum_{l=1}^{M} K(z,y^{(l)})P(y^{(l)})$ for any
polynomial $P$ of  degree $d_y-1$ or less.
For numerical calculations of $K(y,y)$ see Ref. \cite{2015arXiv151005510G}, Appendix C.

The $\lambda(y)$ define a value similar in nature to ``the number of observations'',
or ``probability'', ``weight'', etc\cite{totik,nevai}.
(The equation $M=\sum_{i=0}^{d_y-1}\lambda(y_i)$ holds,  when $y_i$
correspond to quadrature nodes build on $y$--distribution, the $y_i$ are eigenvalues
of generalized eigenfunctions problem:
$\sum_{t=0}^{d_y-1} <yQ_sQ_t>_y\psi_t^{(i)}=y_i \sum_{t=0}^{d_y-1} <Q_sQ_t>_y\psi_t^{(i)}$, also note that $\lambda(y_i)=1/K(y_i,y_i)=1/\left(\sum_{t=0}^{d_y-1}\psi_t^{(i)}Q_t(y_i)\right)^2$ and $0=\sum_{t=0}^{d_y-1}\psi_t^{(s)}Q_t(y_i)$ for $s\ne i$.
The asympthotic of $\lambda(y)$ can also serve as
important characteristics of distribution property\cite{totik}.)
The problem now is to modify $\lambda(y)$ to take into account given $x$ value.
If, in addition to $y^{(l)}$, we have a
vector $x_j^{(l)}$ as precondition, then the weight in (\ref{QQy}) for each $l$,
should be no longer equal to the constant for all terms, but instead,
should be calculated based on the number of
$x_j^{(l)}$ observations that are close to given $x$ value.
Let us use Christoffel function once again,
but now in $x$--space under fixed $l$ and estimate
the weight for $l$-th observation of $y$ as equal to $\lambda^{(l)}(x)$

The result for $\lambda(y|x)$ is:
\begin{eqnarray}
  <Q_k>^{(l)}_x&=&\sum_{j=1}^{N} Q_k(x_j^{(l)}) \label{xmu} \\
  G^{(l)}&=&<Q_qQ_r>^{(l)}_x=\sum_{j=1}^{N} Q_q(x_j^{(l)})Q_r(x_j^{(l)}) \label{QQx} \\
  \lambda^{(l)}(x)&=&\frac{1}{\sum_{k,m=0}^{d_x-1}Q_k(x)\left(G^{(l)}\right)^{-1}_{km}Q_m(x)} \label{lambdal} \\
  <Q_s>_{\lambda}&=&\sum_{l=1}^{M} \lambda^{(l)}(x)Q_s(y^{(l)}) \label{mulambda}\\
  G_{y|x}&=&<Q_sQ_t>_{\lambda}=\sum_{l=1}^{M} \lambda^{(l)}(x)Q_s(y^{(l)})Q_t(y^{(l)}) \label{Gadj}\\
  \lambda(y|x)&=&\frac{1}{\sum_{s,t=0}^{d_y-1}Q_s(y)\left(G_{y|x}\right)^{-1}_{st}Q_t(y)} \label{wxy}
\end{eqnarray}
The $\lambda(y|x)$ is the answer. The $G_{y|x}$ is very similar to (\ref{QQy}),
but now the $l$-th term weight is $\lambda^{(l)}(x)$ instead of a constant.
For a given $x$
 the (\ref{wxy}) is a function of $y$, having the meaning
 of observations number (or ``probability''--like value when scaled).
 The conceptual difference between regressing the value of $y$ on $x$ and
 $x$--dependent weights is conceptually similar to the difference
 between least squares approximation, where observable value is interpolated
 and Radon--Nikodym type of approximation,
 where the weights are interpolated\cite{2015arXiv151101887G}.
 In Christoffel function approach only the weights, not the values
 are interpolated, what gives a new turn to distribution regression problem.

For an estimation of possible
 $y$ outcomes given $x$, this can be done
 either using the (\ref{mulambda}) measure and estimating, say,
 average $y$ and dispersion, or more interesting,
 build $d_y$--point Gauss quadrature using the measure (\ref{mulambda}),
 see Ref. \cite{2015arXiv151005510G}, Appendix B for numerical algorithm,
 and, for the measure (\ref{mulambda}),
 obtain quadrature nodes $y_i$ and weights $\lambda(y_i|x)$.
 Then quadrature nodes $y_i$ can be treated
 as possible $y$--outcomes and $\lambda(y_i|x)$ can be treated as the
 weight, corresponding to $y_i$ outcome. Normalizing the
 weights one receive probabilities of each $y_i$--outcome given $x$ value.
 (The quadratures provide superior information about probabilities
 of each outcome, taking long--tail information into account,
 but if one, for whatever reason, still need average $y$ value,
 corresponding to (\ref{mulambda}) measure,
 it can be easily obtained from quadrature averaging $y_i$
 with probabilities $\lambda(y_i|x)/\sum_{m=0}^{d_y-1}\lambda(y_m|x)$ of $y_i$ outcome.
 The result would match exactly sample average of $y$ for the measure (\ref{mulambda}).
Also note that
 $\sum_{i=0}^{d_y-1}\lambda(y_i|x)=\sum_{l=1}^{M}\lambda^{(l)}(x)$).

\section{\label{christoffelnum}Numerical Estimation}
The major problem of Christoffel function  calculation
is numerical instability.
For given observations all polynomial bases give identical results,
but numerical stability of calculations is drastically different,
because Gramm matrix condition number depend strongly
on basis choice.
If $Q_k(x)$ and $Q_m(y)$ are chosen
as orthogonal polynomials with orthogonality measure support matching the $x$ and $y$ support
then for discrete measures the Gramm matrix  posses
a good condition number\cite{beckermann1996numerical}.
The numerical library we developed, see\cite{2015arXiv151005510G} Appendix A,
is able to manipulate polynomials 
in Chebyshev, Legendre, Laguerre and Hermite bases directly,
what allows a stable basis to be used
and calculate the moments to a very high order, see Ref. \cite{2015arXiv151101887G} as an example.
The distribution regression problem does not require hundreds of moments
as in \cite{2015arXiv151101887G}, the
$d_x$ and $d_y$ are typically lower than 10--15 and also should be
substantially lower that $N$ and $M$ values respectively.
The numerical calculations
are typically stable as long as one of four stable bases from Ref. \cite{2015arXiv151005510G}
is used.

\begin{figure}[t]
\includegraphics[width=10cm]{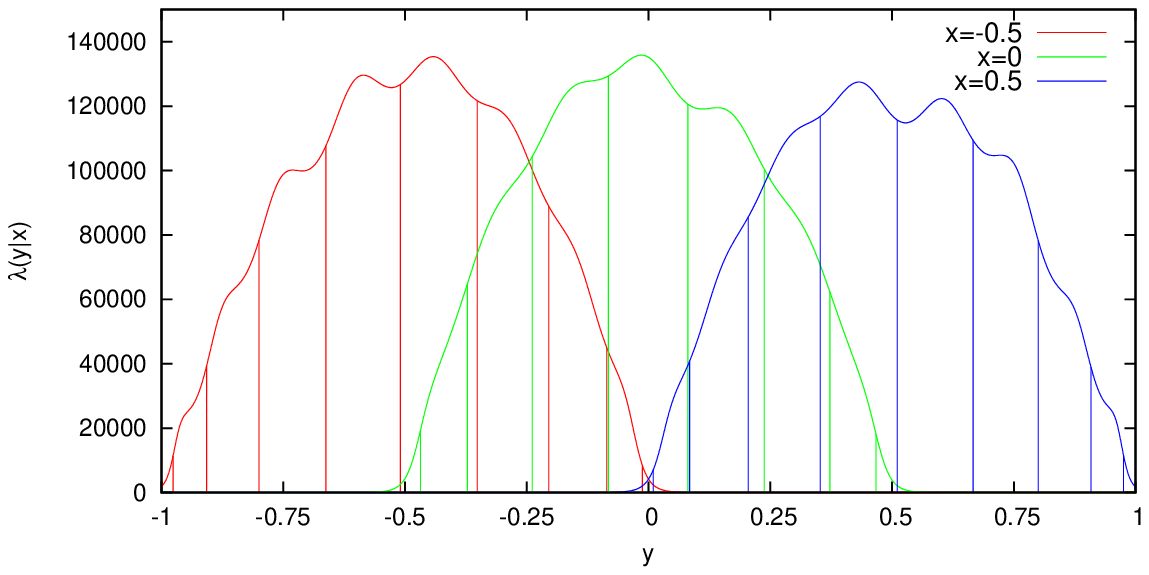}
\includegraphics[width=10cm]{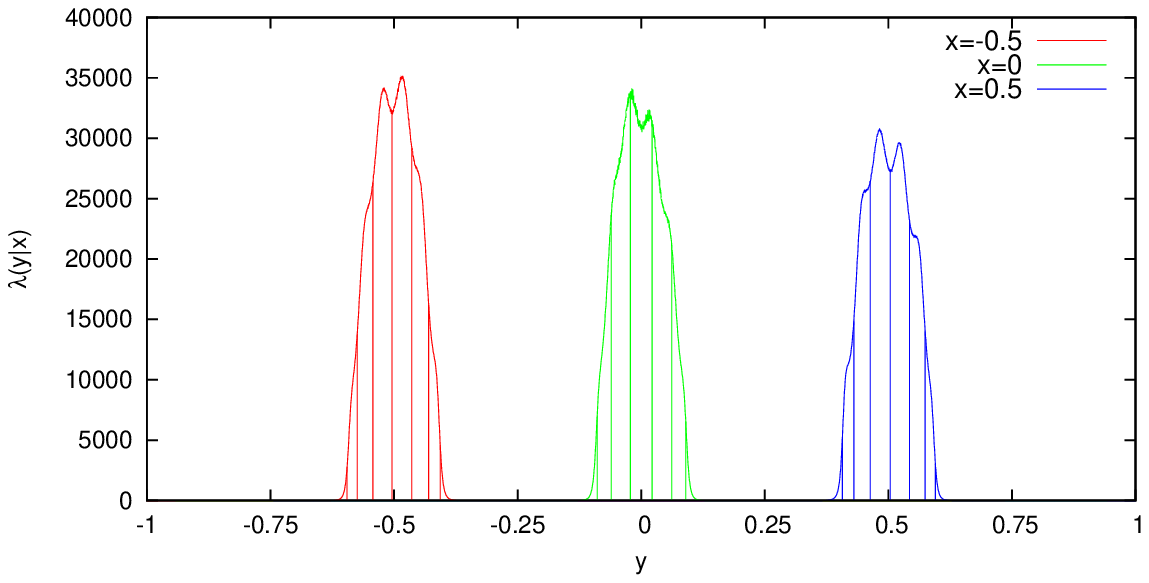}
\includegraphics[width=10cm]{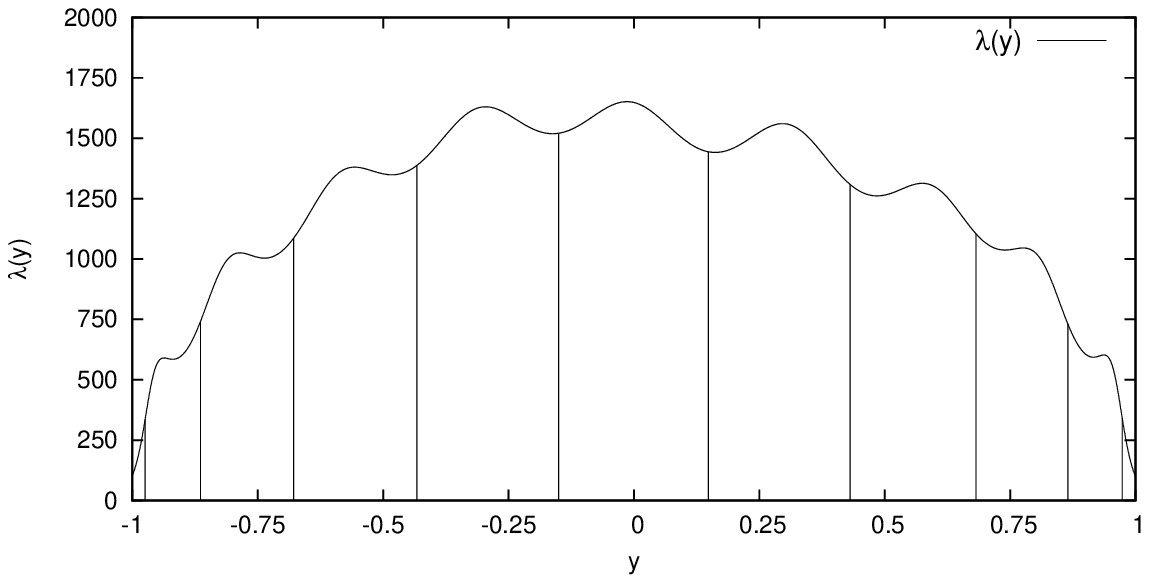}
\caption{\label{fig:R}
  The $\lambda(y|x)$ for $d_x=d_y=10$ and given $x=\{-0.5,0,0.5\}$ for $R=0.5$ (top)
  and $R=0.1$ (middle).
  The $\lambda(y)$ from (\ref{w}) (bottom). To have $\lambda(y)$ 
  similar to $\lambda(y|x)$ scale it should be multiplied by a factor about $N/d_x$.
  The vertical lines correspond to quadrature built on $\lambda$ -- measure,
  each line height is a weight corresponding to specific outcome.
  }
\end{figure}

The algorithm for  $\lambda(y|x)$ calculation is this.
For each $l$ calculate: $<Q_k>^{(l)}_x=\sum_{j=1}^{N} Q_k(x_j^{(l)})$ moments for $k=[0..2d_x-1]$,
then, using polynomials multiplication operation, from these moments
obtain Gramm matrix (\ref{QQx}) $G^{(l)}=<Q_qQ_r>^{(l)}_x$ for $q,r=[0..d_x-1]$,
inverse it and build $\lambda^{(l)}(x)$, a rational function (the nominator is a constant and
the denominator is a polynomial of $2d_x-2$ order) as in (\ref{lambdal}),
then calculate the $\lambda^{(l)}(x)$  at given value of $x$, save these
as the weight for $l$-th observation of $y^{(l)}$.
Having the weights, conditional on given $x$ value,  calculate
$<Q_m>_{\lambda}=\sum_{l=1}^{M} \lambda^{(l)}(x) Q_m(y^{(l)})$ moments for $m=[0..2d_y-1]$, then,
using polynomials multiplication operation, from these moments
obtain Gramm matrix (\ref{Gadj}) $G_{y|x}=<Q_sQ_t>_{\lambda}$ for $s,t=[0..d_y-1]$,
inverse it and build $\lambda(y|x)$ as in (\ref{wxy}).
If possible $y$ outcomes and their probabilities are required, then solve generalized
eigenvalues problem
$\sum_{t=0}^{d_y-1} <yQ_sQ_t>_{\lambda}\psi_t^{(i)}=y_i \sum_{t=0}^{d_y-1} <Q_sQ_t>_{\lambda}\psi_t^{(i)}$, the eigenvalues $y_i$ provide
possible $y$--outcomes and the weight for each outcome
is $\lambda(y_i|x)=1/\left(\sum_{t=0}^{d_y-1}\psi_t^{(i)}Q_t(y_i)\right)^2$,
the probability of $i$--th outcome is normalized weight
$\lambda(y_i|x)/\sum_{m=0}^{d_y-1}\lambda(y_m|x)$
The code performing these calculations is available\cite{polynomialcode},
see the file ExampleDistributionDependence.scala.

For application of the algorithm 
consider the following simple numerical example.
Let $y$ be uniformly $[-1\dots 1]$ distributed random variable, $l=[1..M];M=10000$,
and for each $y^{(l)}$ generate $j=[1..N];N=1000$ random $x$ as $x=y+R*\epsilon[-1\dots 1]$,
where $\epsilon[-1\dots 1]$ is uniformly $[-1\dots 1]$ distributed random variable.
Then for given $x$, we want to estimate the distribution of $y$.
Let us choose $d_x=d_y=10$ and plot $\lambda(y|x)$, the function of $y$
for three fixed $x=\{-0.5,0,0.5\}$.
In the Fig. \ref{fig:R} we present the chart
for $\lambda(y|x)$ for $R=0.5$ and $R=0.1$.
Unconditional $\lambda(y)$ from (\ref{w}) is also presented.
(For some applications conditional $\lambda(y|x)/\lambda(y)$ can be also considered).

One can see that the  $y$--localization at given $x$ is
very clear, and the width of non--vanishing area of $\lambda(y|x)$
track very close the value of randomness parameter $R$.
The quadrature built on $\lambda$ measure give
both: possible outcomes (quadrature nodes) and weights
(outcome probability is normalized weight), presented in the Fig. \ref{fig:R}
as vertical lines corresponding to specific $y$--outcome
(as we noted above -- on quadrature nodes the quadrature weight match
exactly Christoffel function value).
These calculations can be applied to any kind of distribution,
this simple example was used just to demonstrate
application of Christoffel function to representation of learned knowledge
and to find possible $y$--outcomes and their probabilities.

\section{\label{christoffeldisc}Discussion}
In this work a Christoffel function approach
to distribution regression problem is proposed.
The main idea is to use Christoffel function
for knowledge representation. Closed form answer (\ref{wxy}) is available.
The Christoffel function is used twice:
first, to build distribution approximation withing a ``bag'',
then to model $y$--value distribution of these ``bags'' using
Christoffel function value from the first step as the weight
for the observation of $y$.
When required, possible $y$ outcomes and their probabilities,
can be calculated by building Gauss quadrature instead
of using plain Christoffel function answer (\ref{wxy}),
the quadrature nodes give possible outcomes
and normalized quadrature weights give each outcome probability.
The method can be extended from real--value to
discrete attributes (the $d_x$ and $d_y$ should be properly adjusted).

The approach, proposed in this paper, as most Multiple--Instance learning approaches,
has two stages.
The question arise, whether consistent one stage approach exist.
For the case $x$ and $y$ being random variables
two one--stage interpolation approaches: least squares and Radon--Nikodym have been
have been studied in \cite{2015arXiv151101887G}.
Now, let us try to find similar one--stage
approach, but for random distribution to random variable
mapping, same as the (\ref{regressionproblem}) problem, we study in this paper.
The idea is to convert the problem ``random distribution'' to ``random variable''
to the problem ``vector of random variables'' to ``random variable''.
The simplest way is to take the moments of random $x$ distribution
as ``input vector of random variables''.
Then least squares and Radon--Nikodym approximations from \cite{2015arXiv151101887G}
can be directly applied.
In contrast with the problem (\ref{regressionproblem}): given $x$, what can we tell about $y$,
this, converted to vector moments problem, would be:
given $<Q_k>_x$ moments of $x$--distribution (fixed $x_0$ case can be modeled by $NQ_k(x_0)$),
what can we tell about $y$? This problem is solvable in one step.
The one--step solution is actually
almost identical to 2D problem of image grayscale intensity interpolation we have
considered in  \cite{2015arXiv151101887G}.
There is just one major difference: for image interpolation problem we used
basis value at specific point of the raster,
but now we would have to use as input the $<Q_k>_x$ moments of $x$ distribution
on which we want to estimate output $y$.
The question arise of numerical stability of one--stage  method
and the problem of data overfitting.
While two--stages approach typically effectively have $d_x+d_y$ elements in basis,
the one--stage approach  effectively have $d_xd_y$ elements in basis.
By choosing stable basis in \cite{2015arXiv151101887G} we calculated
the moments for $d_x=d_y=100$ without catching an instability,
so basis dimension should not be an issue, but the question
of data overfitting for one--stage method require more research
and to be published separately. The major advantage
of using Christoffel function for knowledge representation is that it stores pure weights,
and data overfitting parameter can be estimated as $d_x/N$ on first stage and about $d_y/M$ on second stage,
so in practical applications the problem can be always identified from the beginning.

\bibliography{LD}

\end{document}